\documentclass{article} 
\usepackage{iclr2026_conference,times}

\iclrfinalcopy

\usepackage{amsmath,amsfonts,bm}

\def\eqref#1{equation~\ref{#1}}

\def\1{\bm{1}}

\DeclareMathAlphabet{\mathsfit}{\encodingdefault}{\sfdefault}{m}{sl}
\SetMathAlphabet{\mathsfit}{bold}{\encodingdefault}{\sfdefault}{bx}{n}

\usepackage{hyperref}
\usepackage{url}
\usepackage{graphicx}

\usepackage{booktabs}   
\usepackage{tabularx}   
\usepackage{adjustbox}  
\usepackage{makecell}   
\usepackage{multirow}

\usepackage{fontspec}
\newfontfamily\figtree[
  Path=fonts/,
  UprightFont=Figtree-Regular.ttf,
  BoldFont=Figtree-Bold.ttf,
  ItalicFont=Figtree-Italic.ttf,
  BoldItalicFont=Figtree-BoldItalic.ttf,
]{Figtree}

\usepackage[boxed]{algorithm}
\usepackage{algpseudocode}
\usepackage{tikz}
\usetikzlibrary{arrows.meta,positioning,calc,fit,shapes.geometric,decorations.pathmorphing}

\colorlet{projgray}{black!10}
\colorlet{panel}{black!5}
\colorlet{bluebox}{blue!25}
\colorlet{greenbox}{green!25}
\colorlet{orangebox}{orange!35}
\colorlet{redbox}{red!30}
\colorlet{linearg}{black!20}

\usepackage{bbding}
\usepackage[table]{xcolor}
\definecolor{ksFrame}{HTML}{111827}
\definecolor{ksBlock}{HTML}{F0F0F0}
\definecolor{ksMain}{HTML}{9355A0}
\definecolor{ksGCN}{HTML}{FF931E}
\definecolor{ksFF}{HTML}{246A3C}
\definecolor{ksGCN2}{HTML}{D73E5F}
\definecolor{ksLinear}{HTML}{1C2127}
\definecolor{ksParam}{HTML}{D6E0FF}
\definecolor{ksSnow}{HTML}{5A61B6}

\tikzset{
  >={Stealth[length=3.5pt]},
  bigbox/.style={draw,rounded corners=16pt,thick,fill=panel,minimum width=6.6cm,minimum height=8.2cm},
  blockbox/.style={draw,rounded corners=20pt,thick,fill=panel,minimum width=8.6cm,minimum height=12.2cm},
  small/.style={draw,rounded corners=8pt,thick,align=center,minimum width=5.2cm,minimum height=8mm},
  call/.style={draw,rounded corners=14pt,thick,minimum width=6.3cm,minimum height=4.1cm,inner sep=10pt},
  proj/.style={draw,thick,shape=trapezium,trapezium angle=64,minimum width=5.0cm,minimum height=1.4cm,align=center,fill=projgray},
  add/.style={draw,thick,circle,minimum size=7.5mm,inner sep=0pt},
}

\title{Task Scarcity and Label Leakage in\\ Relational Transfer Learning}

\author{
Francisco Galuppo Azevedo$^{1,2}$, Clarissa Lima Loures$^{1,2}$, Denis Oliveira Correa$^{1}$ \\
$^{1}$Kunumi Institute, Belo Horizonte, Minas Gerais, Brazil \\
$^{2}$Universidade Federal de Minas Gerais, Belo Horizonte, Minas Gerais, Brazil \\
\texttt{\{francisco,clarissa.loures,denis\}@kunumi.com}
}

\begin{document}

\maketitle

\begin{abstract}
Training relational foundation models requires learning representations that transfer across tasks, yet available supervision is typically limited to a small number of prediction targets per database. This task scarcity causes learned representations to encode task-specific shortcuts that degrade transfer even within the same schema, a problem we call label leakage. We study this using K-Space, a modular architecture combining frozen pretrained tabular encoders with a lightweight message-passing core. To suppress leakage, we introduce a gradient projection method that removes label-predictive directions from representation updates. On RelBench, this improves within-dataset transfer by +0.145 AUROC on average, often recovering near single-task performance. Our results suggest that limited task diversity, not just limited data, constrains relational foundation models.
\end{abstract}

\section{Introduction}
Relational databases remain the dominant substrate for operational data. Building foundation models for this setting requires representations that transfer across tasks. Yet a typical relational database offers only a handful of supervised prediction tasks. This task scarcity is a data bottleneck distinct from sample scarcity: even with millions of rows, the number of distinct objectives available to shape representations is small.

When supervision comes from a single task, learned representations are free to encode any feature that predicts the label, including task-specific shortcuts that do not generalize. We call this label leakage: the representation absorbs label-predictive directions that help the training objective but hurt transfer. With abundant task diversity, conflicting gradients from different objectives would suppress such shortcuts. Under task scarcity, no such pressure exists. As we show empirically, within-dataset transfer can collapse to near-chance performance even when single-task accuracy is strong.

We study this problem within the Relational Deep Learning (RDL) framework, which constructs heterogeneous graphs from multi-table databases and learns directly over that structure \cite{fey2023rdl,robinson2024relbench}. Our testbed is \textbf{K-Space}, a modular RDL architecture with a convolutional message-passing core that has linear edge complexity. Row-level representations come from a frozen pretrained tabular encoder, and predictions are made by a frozen in-context learning head \cite{qu2025tabicl}. This modularity isolates the relational backbone as the component where leakage can occur. To suppress it, we introduce a gradient projection method: at each training step, we remove from the main loss gradient any component that aligns with an adversarial label-prediction gradient, computed per sample at the representation level.

\vspace{0.5em}
\noindent\textbf{Contributions.}
\begin{itemize}
\item We empirically demonstrate that task scarcity causes label leakage in relational learning: within-dataset transfer often collapses to near-chance performance despite strong single-task accuracy.
\item We propose a sample-wise gradient projection method that suppresses label-predictive directions in the shared representation during training.
\item We show that this intervention improves within-dataset transfer (+0.145 AUROC on average) but does not help and can hurt cross-dataset transfer, suggesting different mechanisms underlie the two settings.
\end{itemize}
\section{Related Work}

\paragraph{RDL and RelBench.} Relational deep learning constructs heterogeneous graphs from multi-table databases and trains neural predictors over these structures \cite{fey2023rdl}. RelBench provides standardized datasets and tasks for this setting, with multiple tasks per database \cite{robinson2024relbench}. This enables evaluation of within-dataset transfer.

\paragraph{Relational transformers.} Recent transformer-based approaches target strong single-task performance and transfer \cite{dwivedi2025relgt,wang2025griffin,ranjan2025relationaltransformer}. RelGT achieves the highest reported results on RelBench under supervised training and serves as our primary baseline. Relational Transformer takes a different approach, using masked prediction to create self-supervised tasks from unlabeled data, sidestepping task scarcity. We explore a complementary direction: mitigating label leakage when only supervised tasks are available.

\paragraph{Pretrained tabular encoders and ICL heads.}
TabICL provides a scalable tabular ICL model whose embedding stages yield strong row representations \cite{qu2025tabicl}. TabPFN-style models illustrate that ICL heads can be effective predictors for tabular tasks, including variants that support regression \cite{hollmann2022tabpfn}. Freezing both encoder and predictor isolates the relational backbone as the sole learnable component and the locus of potential label leakage.

\paragraph{Scalable message passing.}
SMPNN shows that deep message passing can be stabilized using transformer-like residual blocks while preserving linear edge complexity \cite{borde2024smpnn}. K-Space adopts SMPNN as the relational building block and extends it to heterogeneous directed propagation with shared weights. A simple architecture allows transfer failures to be attributed to the representation rather than architectural mismatches.

\paragraph{Adversarial methods for representation learning.} Adversarial training can remove unwanted information from representations. Domain-adversarial neural networks (DANN) use gradient reversal to learn domain-invariant features \cite{ganin2015dann}, but we found this unstable for removing task-specific label information. Instead, inspired by Muon \cite{jordan2024muon}, we subtract the adversarial gradient's component from the main gradient.
\section{Architecture}

\begin{figure}[h]
\centering
\begin{tikzpicture}[font=\small\figtree, yscale=0.81, xscale=0.9, transform shape]
\tikzset{
  conn/.style={draw=ksFrame, thick},
  projfill/.style={draw=ksFrame, fill=ksLinear!18, thick},
  blockstyle/.style={draw=ksFrame, fill=ksBlock!80, thick,
                     minimum width=3.4cm, minimum height=3.4cm, align=center},
  miniblock/.style={draw=ksFrame, fill=ksBlock!20, thick,
                    minimum width=1.2cm, minimum height=1.8cm, align=center},
  lane/.style={draw=ksFrame, thick},
  addnode/.style={draw=ksFrame, circle, inner sep=0pt,
                  minimum size=9pt, thick, fill=white},
  contbox/.style={draw=ksFrame, thick, fill=ksBlock!40},
  thinblock/.style={thick, minimum width=3.0cm, minimum height=0.5cm, align=center},
  blk/.style={thinblock, draw=ksMain,  fill=ksMain!10},
  gcnblk/.style={thinblock, draw=ksGCN, fill=ksGCN!12},
  ffblk/.style={thinblock, draw=ksFF,  fill=ksFF!12},
  zoom/.style={draw=ksFrame!60, thick, dashed},
  gcnblk2/.style={thinblock, draw=ksGCN2, fill=ksGCN2!18},
  linear/.style={thinblock, draw=ksLinear, fill=ksLinear!18},
}


\begin{scope}[shift={(0,0.3)}]
  \coordinate (inputtop) at (0,0.5);
  \path[projfill]
    (-1.7,0.5) -- (1.7,0.5) -- (1.2,-0.5) -- (-1.2,-0.5) -- cycle;
  \node at (0,0) {Input Projection};
\end{scope}

\coordinate (inputbot) at (0,-0.2);

\node[blockstyle] (smpnn) at (0,3.3) {%
  \begin{minipage}[t][3cm][t]{2.8cm}
    \centering
    Hetero Block
  \end{minipage}
};

\node at (-2.2,3.2) {\Large $N\times$};

\node[miniblock] (rev) at (-0.9,2.9) {REV\\Block};
\node[miniblock] (fwd) at (0.9,2.9) {FWD\\Block};

\coordinate (hetero_in)  at (0,1.6);
\coordinate (hetero_out) at (0,5.0);

\coordinate (hetero_start) at (0,1.7);
\coordinate (hetero_split) at (0,1.85);
\coordinate (hetero_end)   at (0,4.5);

\node[addnode] (hetero_add) at (0,4.0) {$+$};

\draw[conn]    (hetero_start) -- (hetero_split);
\draw[conn]    (hetero_split) -- (hetero_add.south);
\draw[conn,->] (hetero_add.north) -- (hetero_end);

\draw[conn] (hetero_split) -| (rev.south);
\draw[conn] (rev.north)   |- (hetero_add.west);

\draw[conn] (hetero_split) -| (fwd.south);
\draw[conn] (fwd.north)   |- (hetero_add.east);

\node[blk, minimum width=3.4cm] (rope) at (0,1.2) {RoPE};

\draw[conn] (inputtop) -- (rope.south);
\draw[conn] (rope.north) -- (hetero_in);

\begin{scope}[shift={(0,5.7)}]
  \coordinate (outbot) at (0,-0.5);
  \coordinate (outtop) at (0,0.5);
  \path[projfill]
    (-1.2,0.5) -- (1.2,0.5) -- (1.7,-0.5) -- (-1.7,-0.5) -- cycle;
  \node at (0,0) {Output projection};
\end{scope}

\draw[conn] (hetero_out) -- (0,5.2);


\begin{scope}[shift={(4.4,0.0)}]

  \path[contbox] (-2.0,0) rectangle (1.8,5.5);

  \node[blk]     (ln1)    at (0,0.8)  {Layer Norm};
  \node[gcnblk]  (gcn1)   at (0,1.4)  {GCN};
  \node[blk]     (scale1) at (0,2.0)  {Scale};
  \node[addnode] (add1)   at (0,2.6)  {$+$};

  \node[blk]     (ln2)    at (0,3.2)  {Layer Norm};
  \node[ffblk]   (ff2)    at (0,4.0)  {Pointwise\\Feedforward};
  \node[blk]     (scale2) at (0,4.8)  {Scale};

  \coordinate (mid_in) at (0,0.1);
  \draw[lane] (mid_in) -- ++(0,0.2) coordinate (splitA);
  \draw[lane] (splitA)       -- (ln1.south);
  \draw[lane] (ln1.north)    -- (gcn1.south);
  \draw[lane] (gcn1.north)   -- (scale1.south);
  \draw[lane] (scale1.north) -- (add1.south);

  \coordinate (railA) at (-1.7,2.6);
  \draw[lane] (splitA) -| (railA) -- (add1.west);

  \draw[lane] (add1.north)  -- (ln2.south);
  \draw[lane] (ln2.north)   -- (ff2.south);
  \draw[lane] (ff2.north)   -- (scale2.south);
  \draw[lane,->] (scale2.north) -- ++(0,0.3);

  \node at (-0.15,5.7) {\normalsize Modified SMPNN Block};

\end{scope}

\draw[zoom, draw=gray] (1.5,2.0) -- (2.4,0.0);
\draw[zoom, draw=gray] (1.5,3.8) -- (2.4,5.5);


\begin{scope}[shift={(8.4,0.0)}]

  \node at (-1.4,2.8) {\normalsize GCN};
  \path[gcnblk] (-1.8,0.2) rectangle (1.8,2.6);

  \node[linear]  (ln1)    at (0,0.8)  {Linear Projection};
  \node[gcnblk2] (gcn1)   at (0,1.4)  {GCNConv / GATv2};
  \node[blk]     (scale1) at (0,2.0)  {SiLU};

  \coordinate (mid_in) at (0.0,0.3);
  \draw[lane]    (mid_in)       -- (ln1.south);
  \draw[lane]    (ln1.north)    -- (gcn1.south);
  \draw[lane]    (gcn1.north)   -- (scale1.south);
  \draw[lane,->] (scale1.north) -- ++(0,0.25);

  \node at (-0.8,5.5) {\normalsize Pointwise FF};
  \path[ffblk] (-1.8,3.5) rectangle (1.8,5.3);

  \node[linear] (ln2)    at (0,4.1) {Linear Projection};
  \node[blk]    (scale2) at (0,4.7) {SiLU};

  \coordinate (mid_in_2) at (0.0,3.6);
  \draw[lane]    (mid_in_2)     -- (ln2.south);
  \draw[lane]    (ln2.north)    -- (scale2.south);
  \draw[lane,->] (scale2.north) -- ++(0,0.25);

\end{scope}

\draw[zoom, draw=ksGCN] (5.9,1.15) -- (6.6,0.2);
\draw[zoom, draw=ksGCN] (5.9,1.65) -- (6.6,2.6);

\draw[zoom, draw=ksFF] (5.9,3.6) -- (6.6,3.5);
\draw[zoom, draw=ksFF] (5.9,4.42) -- (6.6,5.3);


\def\figXmin{-1.6}
\def\figXmax{10.2}
\pgfmathsetmacro{\myx}{(\figXmax-2.8)/2 + 2.8}

\def\topYmin{6.6}
\def\topYmax{7.1}

\path[draw=ksLinear, line width=0.5pt, fill=ksLinear!18]
  (\figXmin,\topYmin) rectangle (1.6,\topYmax);
\node at (0.0,6.85) {Adversarial Head};

\path[draw=ksParam, line width=0.5pt, fill=ksParam!80]
  (2.8,\topYmin) rectangle (\figXmax,\topYmax);

\coordinate (topPredL) at (2.8+0.3,6.85);
\coordinate (topPredR) at (\figXmax-0.3,6.85);

\node[anchor=west] at (topPredL) {[TabICL] Predictor};
\node[anchor=east] at (topPredR) {25.78M \phantom{aaa}};
\node[anchor=east] at (topPredR) {\color{ksSnow}\SnowflakeChevron};

\coordinate (out_split) at (0,6.4);
\coordinate (adv_in)    at (0.0,\topYmin);
\coordinate (pred_in)   at (\myx,\topYmin);

\coordinate (adv_knee)  at (0.0,6.4);
\coordinate (pred_knee) at (\myx,6.4);

\draw[conn] (outtop) -- (out_split);

\draw[conn]    (out_split) -- (0,6.2);
\draw[conn] (0,6.4) -- (adv_knee) -- (adv_in);

\draw[conn]    (out_split) -- (0,6.4);
\draw[conn] (0,6.4) -- (pred_knee) -- (pred_in);

\coordinate (adv_head_top)  at (0.0,\topYmax);
\coordinate (pred_head_top) at (\myx,\topYmax);

\coordinate (adv_arrow_end)  at (0.0,7.4);
\coordinate (pred_arrow_end) at (\myx,7.4);

\draw[conn,->] (adv_head_top)  -- (adv_arrow_end);
\draw[conn,->] (pred_head_top) -- (pred_arrow_end);

\node[anchor=south] at (adv_arrow_end)  {Label Prediction};
\node[anchor=south] at (pred_arrow_end) {In-Context Label Prediction};


\def\botYmin{-1.4}
\def\botYmax{-0.9}
\def\tabFeatYmin{-2.2}
\def\tabFeatYmax{-1.7}

\node at (-1.3,-1.1) {Time Features};
\node at (1.2,-1.1) {RWPE};

\node at (\myx,-1.9) {Table Features};

\path[draw=ksParam, line width=0.5pt, fill=ksParam!80]
  (2.8,\botYmin) rectangle (\figXmax,\botYmax);

\coordinate (botMainL) at (3.1,-1.15);
\coordinate (botMainR) at (\figXmax-0.3,-1.15);

\node[anchor=west] at (botMainL)
  {[TabICL] Encoder};
\node[anchor=east] at (botMainR) {1.28M \phantom{aaa}};
\node[anchor=east] at (botMainR) {\color{ksSnow}\SnowflakeChevron};

\coordinate (tabfeat_out)  at (\myx,\tabFeatYmax);
\coordinate (time_out)     at (-1.3,\botYmax);
\coordinate (rwpe_out)     at ( 1.2,\botYmax);
\coordinate (tabicl_out)   at ( \myx,\botYmax);
\coordinate (tabicl_in)    at ( \myx,\botYmin);
\coordinate (in_split)     at (0,-0.6);
\coordinate (time_knee)    at (-1.3,-0.6);
\coordinate (rwpe_knee)    at ( 1.2,-0.6);
\coordinate (tabicl_knee)  at ( \myx,-0.6);

\draw[conn]   (time_out)   -- (time_knee)   -- (in_split);
\draw[conn]   (rwpe_out)   -- (rwpe_knee)   -- (in_split);
\draw[conn]   (tabicl_out) -- (tabicl_knee) -- (in_split);
\draw[conn]   (tabfeat_out) -- (tabicl_in);
\draw[conn,->](in_split) -- (inputbot);

\end{tikzpicture}
\caption{Schematic of the proposed architecture. A TabICL encoder (column embedder and row interactor) encodes table features; its output is concatenated with RWPE and time features and passed through an input projection and per-table-type RoPE into a stack of $N$ Hetero (SMPNN) blocks with reversible (REV) and forward (FWD) paths. Zoomed views show the SMPNN block and its GCN and pointwise feedforward sub-blocks. An output projection feeds two heads: an adversarial MLP head and a TabICL predictor head.}
\label{fig:arch}
\end{figure}
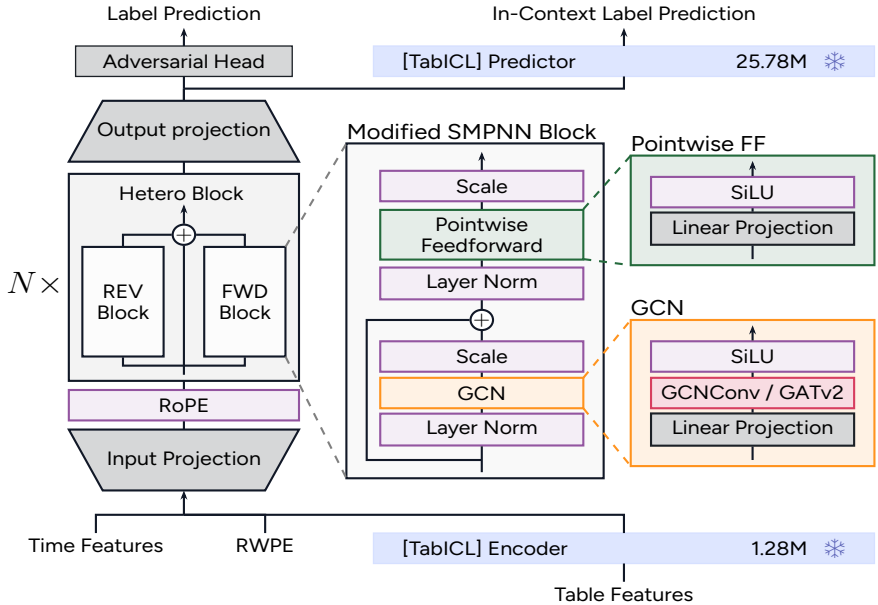

We now describe the K-Space architecture and our gradient projection implementation. It has three blocks: a frozen table encoder, a relational convolutional core, and an ICL predictor head, see Figure \ref{fig:arch}.

\paragraph{Frozen table encoder.} Each node begins as a row in some table. We compute row embeddings using the frozen embedding stages of TabICL \cite{qu2025tabicl}. To these, we concatenate temporal features (index plus periodic encodings) and random walk positional encodings (RWPE). Unlike the original RWPE formulation \cite{dwivedi2021lspe}, which uses matrix exponentiation, we compute encodings via explicit random walks that cannot traverse future edges. The result is projected to the model width, with table type injected via RoPE \cite{su2021roformer}.

\paragraph{Relational convolution.} We build on SMPNN \cite{borde2024smpnn}. Each layer uses shared parameters across relation types and separates incoming from outgoing propagation. For each relation type r, we compute an update from incoming neighbors and an update from outgoing neighbors, aggregate across r, and apply a residual update. The full procedure is given in Algorithm~\ref{alg} (Appendix~\ref{app:architecture}).

\paragraph{ICL predictor head}
We predict binary labels via a pretrained frozen ICL head: each mini-batch splits labeled instances into support (embedding, label) pairs and query embeddings. The head consumes support context and outputs query probabilities. Our experiments focus on classification due to the TabICL predictor interface \cite{qu2025tabicl}.

\paragraph{Label-leakage suppression}
All reported K-Space results use the same architecture; we vary only whether training includes an adversarial label-prediction head. The motivation is to reduce direct label information in representations, which can hinder transfer when tasks change. 

\begin{figure}[t]
  \centering
\begin{tikzpicture}[font=\footnotesize,>=stealth]
  \tikzset{
    axis/.style={->, line width=1.5pt, draw=black!10},
    sample/.style={circle, fill=black, inner sep=2pt},
    gradICL/.style={->, line width=1.8pt, color=green!60!black},
    gradAdv/.style={->, line width=1.8pt, color=red!65!black},
    gradICLRes/.style={->, line width=1.6pt, color=green!40!black},
    projLine/.style={dashed, line width=1.2pt, draw=gray!60}
  }

  \begin{scope}[yshift=0.7cm]
    \draw[axis] (0.6,2) -- (6.2,2);
    \draw[axis] (2.8,0.6) -- (2.8,4.0);

    \coordinate (x1) at (2.1,1.3);
    \coordinate (x2) at (3.3,3.0);
    \coordinate (x3) at (4.5,1.4);

    \coordinate (x1AdvTip) at ($(x1)+(0.55,0.70)$);
    \coordinate (x2AdvTip) at ($(x2)+(0.25,0.55)$);
    \coordinate (x3AdvTip) at ($(x3)+(0.85,0.80)$);

    \coordinate (x1IclTip) at ($(x1)+(-0.50,1.00)$);
    \coordinate (x2IclTip) at ($(x2)+(-0.60,0.20)$); 
    \coordinate (x3IclTip) at ($(x3)+(-0.40,0.90)$);

    \coordinate (x1Proj) at ($(x1)+(0.2950,0.3754)$);
    \coordinate (x3Proj) at ($(x3)+(0.2371,0.2231)$);

    \draw[projLine] (x1IclTip) -- (x1Proj);
    \draw[projLine] (x3IclTip) -- (x3Proj);

    \draw[gradAdv] (x1) -- (x1AdvTip);
    \draw[gradAdv] (x2) -- (x2AdvTip);
    \draw[gradAdv] (x3) -- (x3AdvTip);

    \draw[gradICL] (x1) -- (x1IclTip);
    \draw[gradICL] (x2) -- (x2IclTip);
    \draw[gradICL] (x3) -- (x3IclTip);

    \coordinate (x1ResTip) at ($(x1)+(-0.7950,0.6246)$);
    \coordinate (x2ResTip) at (x2IclTip); 
    \coordinate (x3ResTip) at ($(x3)+(-0.6371,0.6769)$);

    \draw[gradICLRes] (x1) -- (x1ResTip);
    \draw[gradICLRes] (x2) -- (x2ResTip);
    \draw[gradICLRes] (x3) -- (x3ResTip);

    \node[sample] at (x1) {};
    \node[sample] at (x2) {};
    \node[sample] at (x3) {};

    \node[anchor=north east] at ($(x1)+(-0.05,-0.05)$) {$x_1$};
    \node[anchor=south east] at ($(x2)+(0.05,-0.5)$)  {$x_2$};
    \node[anchor=north west] at ($(x3)+(0.05,-0.05)$)  {$x_3$};
  \end{scope}

  \node[anchor=south west, align=left, font=\scriptsize] at (6.2,2.4) {%
    \begin{minipage}{0.4125\linewidth}
    \begin{algorithmic}
    \If{$\big\langle \nabla_h \ell_{\mathrm{ICL}}(x_i),
                 \nabla_h \ell_{\mathrm{adv}}(x_i) \big\rangle > 0$}
      \State $\alpha \gets
      \dfrac{\big\langle \nabla_h \ell_{\mathrm{ICL}}(x_i),
                      \nabla_h \ell_{\mathrm{adv}}(x_i) \big\rangle}
             {\big\| \nabla_h \ell_{\mathrm{adv}}(x_i) \big\|_2^2}$
      \State $\tilde{\nabla}_h \ell_{\mathrm{ICL}}(x_i)
      \gets \nabla_h \ell_{\mathrm{ICL}}(x_i)
            - \alpha \,\nabla_h \ell_{\mathrm{adv}}(x_i)$
    \Else
      \State $\tilde{\nabla}_h \ell_{\mathrm{ICL}}(x_i)
      \gets \nabla_h \ell_{\mathrm{ICL}}(x_i)$
    \EndIf
    \end{algorithmic}
    \end{minipage}
  };

  \draw[gradICL]    (7.8,2) -- ++(0.9,0);
  \node[anchor=west] at (8.8,2)
    {$\nabla_h \ell_{\text{ICL}}(x_i)$};

  \draw[gradAdv]    (7.8,1.7) -- ++(0.9,0);
  \node[anchor=west] at (8.8,1.7)
    {$\nabla_h \ell_{\text{adv}}(x_i)$};

  \draw[gradICLRes] (7.8,1.4) -- ++(0.9,0);
  \node[anchor=west] at (8.8,1.4)
    {$\tilde{\nabla}_h \ell_{\text{ICL}}(x_i)$};
\end{tikzpicture}
\caption{Geometry of per-sample gradients in the representation space of $h$, the shared representation just before the two heads. For each sample $x_i$, the light green arrow is the in-context learning gradient $\nabla_h \ell_{\text{ICL}}(x_i)$, the red arrow is the adversary gradient $\nabla_h \ell_{\text{adv}}(x_i)$, and the dark green arrow is the refined gradient $\tilde{\nabla}_h \ell_{\text{ICL}}(x_i)$ after subtracting its projection onto $\nabla_h \ell_{\text{adv}}(x_i)$ (dashed gray segment).}
\label{fig:grad-geometry}
\end{figure}
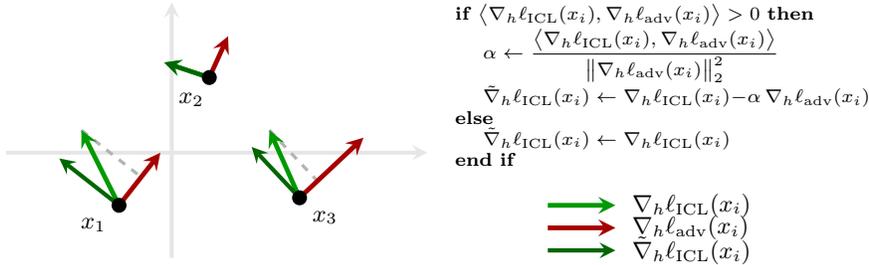

Let $\mathbf{z}$ denote the query representation after the GNN and before the ICL head. We train with the main ICL loss $\ell_{\text{main}}$ on query predictions. We optionally add an adversarial head trained to predict the label from $\mathbf{z}$ alone. When enabled, we compute per-query gradients w.r.t.\ $\mathbf{z}$ and remove components of $\nabla_{\mathbf{z}}\ell_{\text{main}}$ that align with $\nabla_{\mathbf{z}}\ell_{\text{adv}}$ via sample-wise projection, as in Figure \ref{fig:grad-geometry}. For efficiency, we apply this transformation at the representation level and backpropagate the averaged transformed gradient through the backbone.

\section{Experiments}
\label{sec:exp}

We ask two questions. First, does task scarcity cause label leakage that harms transfer? Second, does gradient projection reduce it? We evaluate on binary classification tasks from RelBench \cite{robinson2024relbench} and report AUROC. We compare against RelGT \cite{dwivedi2025relgt}, a strong transformer-based baseline.

To isolate the effect of task scarcity, we evaluate under four supervision regimes:
\begin{itemize}
\item \textbf{ST (Single-task):} Train and evaluate on the same task. This is the standard setting and our upper bound.
\item \textbf{WD (Within-dataset):} Train on a different task from the same database. Same schema, different labels. This is our main test for label leakage.
\item \textbf{CD (Cross-dataset):} Train on all tasks except those from the target database. Different schema entirely.
\item \textbf{ALL (All-task):} Train on all tasks jointly. Tests multi-task interference.
\end{itemize}
Two datasets (\texttt{rel-hm} and \texttt{rel-trial}) have a single task, so WD cannot be computed for these. Additionally, \texttt{rel-event} was excluded from ALL due to memory constraints, and ALL was trained for only 3 epochs (versus up to 20 for other regimes) due to time constraints, so ALL results may underestimate the potential of multi-task training.

For each regime, we report two K-Space variants: \textbf{Base} (without adversarial head) and \textbf{Adv.} (with adversarial head and gradient projection). Both use identical architecture; only the training objective differs. Hyperparameter selection is described in Appendix~\ref{app:hyperparams}. 

\section{Results}
The main results are shown in Table~\ref{tab:main}.

\begin{table}[h]
\centering
\small
\setlength{\tabcolsep}{3pt}
\renewcommand{\arraystretch}{1.08}
\begin{adjustbox}{width=\linewidth}
\begin{tabular}{ll c @{\hspace{6pt}} cc | cc | cc | cc}
\toprule
& & \multicolumn{1}{c}{RelGT} & \multicolumn{8}{c}{K-Space} \\
\cmidrule(lr){3-3}\cmidrule(lr){4-11}
Dataset & Task
& \multicolumn{3}{c}{ST}
& \multicolumn{2}{c}{WD}
& \multicolumn{2}{c}{CD}
& \multicolumn{2}{c}{ALL} \\
\cmidrule(lr){3-5}\cmidrule(lr){6-7}\cmidrule(lr){8-9}\cmidrule(lr){10-11}
& & &
\makecell[c]{Base} & \makecell[c]{Adv.} &
\makecell[c]{Base} & \makecell[c]{Adv.} &
\makecell[c]{Base} & \makecell[c]{Adv.} &
\makecell[c]{Base} & \makecell[c]{Adv.} \\
\midrule

\multirow{2}{*}{rel-f1} & driver-dnf
& 0.759
& 0.735 & 0.729
& 0.368 & 0.702
& 0.645 & 0.447
& 0.609 & 0.446 \\
& driver-top3
& 0.835
& 0.818 & 0.647
& 0.219 & 0.829
& 0.724 & 0.683
& 0.528 & 0.790 \\
\midrule

\multirow{2}{*}{rel-avito} & user-visits
& 0.668
& 0.645 & 0.646
& 0.401 & 0.602
& 0.517 & 0.421
& 0.583 & 0.523 \\
& user-clicks
& 0.683
& 0.646 & 0.646
& 0.378 & 0.635
& 0.577 & 0.416
& 0.572 & 0.536 \\
\midrule

\multirow{2}{*}{rel-event} & user-repeat
& 0.761
& 0.735 & 0.736
& 0.298 & 0.264
& --- & ---
& --- & --- \\
& user-ignore
& 0.816
& 0.852 & 0.858
& 0.498 & 0.504
& --- & ---
& --- & --- \\
\midrule

\multirow{2}{*}{rel-stack} & user-engagement
& 0.905
& 0.836 & 0.844
& 0.809 & 0.799
& 0.647 & 0.739
& 0.818 & 0.818 \\
& user-badge
& 0.863
& 0.833 & 0.833
& 0.501 & 0.610
& 0.751 & 0.698
& 0.845 & 0.835 \\
\midrule

\multirow{2}{*}{rel-amazon} & user-churn
& 0.704
& 0.642 & 0.644
& 0.631 & 0.626
& 0.624 & 0.513
& 0.636 & 0.633 \\
& item-churn
& 0.826
& 0.770 & 0.778
& 0.755 & 0.740
& 0.549 & 0.511
& 0.766 & 0.773 \\
\midrule

rel-hm & user-churn
& 0.693
& 0.674 & 0.673
& --- & ---
& 0.546 & 0.496
& 0.602 & 0.622 \\
\midrule

rel-trial & study-outcome
& 0.686
& 0.651 & 0.627
& --- & ---
& 0.528 & 0.515
& 0.523 & 0.528 \\
\midrule

\multicolumn{2}{l}{Average}
& 0.767
& 0.737 & 0.722
& 0.486 & 0.631
& 0.611 & 0.544
& 0.648 & 0.650 \\
\bottomrule
\end{tabular}
\end{adjustbox}
\caption{\textbf{RelBench AUROC.} ST, WD, CD, ALL are the supervision regimes. Within each regime, the left column is the base model, and the right column ({\scriptsize Adv.}) uses the adversarial head with sample-wise gradient projection. RelGT is the baseline.}
\label{tab:main}
\end{table}

\paragraph{Single-task.}
Under single-task supervision (ST), K-Space Base achieves an average AUROC of 0.737, compared to 0.767 for RelGT. The gap is expected given our modular constraints: frozen encoder, frozen predictor, and a simple message-passing backbone. Despite this, K-Space matches or exceeds RelGT on some tasks (e.g., rel-event/user-ignore: 0.852 vs.\ 0.816). The adversarial variant slightly underperforms Base in ST (0.722 vs.\ 0.737), suggesting that removing label-predictive directions can reduce accuracy when transfer is not required.

\paragraph{Within-dataset.}
Within-dataset transfer (WD) is where gradient projection has the largest effect. Average AUROC increases from 0.486 (Base) to 0.631 (Adv.), a +0.145 improvement. In several cases, Base collapses toward random while Adv.\ recovers to near-ST performance (e.g., rel-f1/driver-top3: 0.219 → 0.829). This supports our hypothesis that the backbone encodes task-specific shortcuts that do not transfer, and that projection substantially reduces this failure mode.

\paragraph{Cross-dataset.}
In cross-dataset transfer (CD), the adversarial variant performs worse than Base (0.544 vs.\ 0.611). One interpretation is that projected-out directions include features useful for cross-schema generalization. Another is that projection reduces effective gradient signal in settings where the model already struggles. Either way, the current method improves task transfer within a dataset but not dataset transfer across schemas.

\paragraph{Multi-task.}
Training on all tasks (ALL) yields similar performance for both variants (Base: 0.648, Adv.: 0.650) and underperforms ST on average. Per-task outcomes vary substantially, consistent with negative transfer under heterogeneous supervision.
\section{Discussion}
Our experiments reveal a clear pattern: under task scarcity, learned representations encode label-predictive shortcuts that do not transfer. Within-dataset transfer collapses to near-chance for the base model, despite strong single-task performance. Gradient projection recovers much of this gap, improving within-dataset transfer by +0.145 AUROC on average. However, the intervention does not help cross-dataset transfer and may even hurt it. Within-dataset and cross-dataset transfer fail differently. The former is dominated by task-specific leakage, while the latter likely involves schema and domain mismatch that our method does not address.

These findings have implications for relational foundation models. Task diversity during training matters more than expected. Self-supervised objectives, such as the masked prediction used by Relational Transformer, offer one way to create abundant tasks from unlabeled data. Explicit leakage suppression offers an alternative approach when training is limited to supervised tasks.

Our study has limitations that suggest future directions: results are classification-only (a regression head such as TabPFN \cite{hollmann2022tabpfn} could be substituted), cross-dataset transfer remains weak (requiring methods beyond gradient projection), and we only evaluate on RelBench. More broadly, understanding when task scarcity harms relational representations remains open.

\bibliographystyle{iclr2026_conference}
\bibliography{iclr2026_conference}

\appendix
\appendix
\section{Architecture Details}
\label{app:architecture}

\begin{algorithm}[H]
\caption{Forward Pass of Hetero Block}
\label{alg}
\begin{algorithmic}[1]
\Require Node features $x$, edge index dictionary $\mathcal{E}$
\State Initialize aggregate vector $A \gets 0$ (same shape as $x$)
\State Initialize count vector $C \gets 0$

\For{each edge type $t$ in $\mathcal{E}$}
    \State Retrieve edge index $E \gets \mathcal{E}[t]$
    \State Determine destination nodes $D \gets \mathrm{unique}(E_{\text{dst}})$
    \State Update count: $C \gets C + \mathbf{1}_{D}$
    \State Select block:
\State $B \gets 
\begin{cases}
\text{block}_{\text{rev}}, & \text{if } t \text{ is reversed} \\
\text{block}_{\text{orig}}, & \text{otherwise}
\end{cases}$
    \State Compute message: $\mathrm{msg} \gets B(x, E)$
    \State Accumulate: $A \gets A + \mathrm{msg} \odot \mathbf{1}_{D}$
\EndFor

\State \Return $x + \dfrac{A}{\max(C, 1)}$
\end{algorithmic}
\end{algorithm}

\section{Hyperparameter Search Details}
\label{app:hyperparams}

We performed sequential hyperparameter optimization to determine the optimal architecture configuration. At each stage, we tuned one hyperparameter while fixing all previously optimized values. We optimized six parameters in order: (1) network depth (number of graph convolutional layers), (2) hidden dimension size for all convolutional layers, (3) neighborhood sampling strategy (number of neighbors sampled at each layer), (4) number of multi-head attention heads, (5) inclusion of temporal features, and (6) dimension of Random Walk Positional Encoding (RWPE) features. 

Table~\ref{tab:hyperparam_overview} summarizes the optimization process across these six stages and results for each hyperparameter search are provided in sequence. We evaluated each configuration on two datasets: \texttt{rel-f1-driver-dnf} and \texttt{rel-avito-user-clicks}, in the WD setting with gradient projection, reporting average AUROC scores. The sequential search improved performance from 0.470 to 0.669.

\begin{table}[h]
\centering
\caption{Sequential Hyperparameter Optimization Overview}
\label{tab:hyperparam_overview}
\begin{tabular}{@{}lllc@{}}
\toprule
Stage & Parameter & Optimal Value & Avg F1 Score \\ 
\midrule
1 & Number of Layers & 3 & 0.581 \\
2 & Hidden Dimension & 256 & 0.590 \\
3 & Neighborhood Size & [16, 8, 4] & 0.611 \\
4 & Attention Heads & 2 & 0.623 \\
5 & Temporal Encoding & Enabled & 0.625 \\
6 & RWPE & 32 & 0.669 \\
\bottomrule
\end{tabular}
\end{table}

\begin{table}[h]
\centering
\caption{Effect of Layer Depth on Model Performance}
\label{tab:layers}
\begin{tabular}{@{}lccccccc@{}}
\toprule
Dataset & \multicolumn{6}{c}{Number of Layers} & \\
\cmidrule(lr){2-7}
 & 1 & 2 & 3 & 4 & 5 & 6 \\ 
\midrule
rel-f1-driver-dnf & 0.519 & 0.588 & {0.587} & 0.527 & 0.541 & 0.553 \\
rel-avito-user-clicks & 0.420 & 0.427 & {0.575} & 0.575 & 0.428 & 0.577 \\
\midrule
Average & 0.470 & 0.507 & \textbf{0.581} & 0.551 & 0.485 & 0.565 \\
\bottomrule
\end{tabular}
\end{table}

\begin{table}[h]
\centering
\caption{Effect of Hidden Dimension on Model Performance}
\label{tab:hidden_dim}
\begin{tabular}{@{}lcccc@{}}
\toprule
Dataset & \multicolumn{3}{c}{Hidden Dimension} & \\
\cmidrule(lr){2-4}
 & 128 & 256 & 512 \\ 
\midrule
rel-f1-driver-dnf & 0.587 & 0.584 & 0.586 \\
rel-avito-user-clicks & 0.575 & {0.597} & 0.587 \\
\midrule
Average & 0.581 & \textbf{0.590} & 0.587 \\
\bottomrule
\end{tabular}
\end{table}

\begin{table}[h]
\centering
\caption{Effect of Neighborhood Configuration on Model Performance}
\label{tab:neighborhood}
\begin{tabular}{@{}lcccc@{}}
\toprule
Dataset & \multicolumn{3}{c}{Neighborhood Configuration} & \\
\cmidrule(lr){2-4}
 & [32, 8, 2] & [16, 8, 4] & [8, 8, 8] \\ 
\midrule
rel-f1-driver-dnf & 0.622 & {0.628} & 0.584 \\
rel-avito-user-clicks & 0.583 & 0.594 & 0.597 \\
\midrule
Average & 0.602 & \textbf{0.611} & 0.590 \\
\bottomrule
\end{tabular}
\end{table}

\begin{table}[h]
\centering
\caption{Effect of Attention Heads on Model Performance}
\label{tab:attention_heads}
\begin{tabular}{@{}lcccc@{}}
\toprule
Dataset & \multicolumn{3}{c}{Number of Attention Heads} & \\
\cmidrule(lr){2-4}
 & 0 & 1 & 2 \\ 
\midrule
rel-f1-driver-dnf & 0.628 & 0.547 & 0.623 \\
rel-avito-user-clicks & 0.594 & 0.621 & {0.624} \\
\midrule
Average & 0.611 & 0.584 & \textbf{0.623} \\
\bottomrule
\end{tabular}
\end{table}

\begin{table}[h]
\centering
\caption{Effect of Temporal Encoding on Model Performance}
\label{tab:temporal}
\begin{tabular}{@{}lccc@{}}
\toprule
Dataset & \multicolumn{2}{c}{Temporal Encoding} & \\
\cmidrule(lr){2-3}
 & Disabled & Enabled \\ 
\midrule
rel-f1-driver-dnf & 0.623 & 0.628 \\
rel-avito-user-clicks & 0.624 & 0.621 \\
\midrule
Average & 0.623 & \textbf{0.625} \\
\bottomrule
\end{tabular}
\end{table}

\begin{table}[h]
\centering
\caption{Effect of Random Walk Positional Encoding (RWPE) on Model Performance}
\label{tab:rwpe}
\begin{tabular}{@{}lccc@{}}
\toprule
Dataset & \multicolumn{2}{c}{RWPE Dimension} & \\
\cmidrule(lr){2-3}
 & 0 (Disabled) & 32 \\ 
\midrule
rel-f1-driver-dnf & 0.628 & 0.702 \\
rel-avito-user-clicks & 0.621 & {0.635} \\
\midrule
Average & 0.625 & \textbf{0.669} \\
\bottomrule
\end{tabular}
\end{table}

\end{document}